\documentclass[11pt]{article}

\usepackage[preprint]{acl}

\usepackage{times}
\usepackage{latexsym}

\usepackage[T1]{fontenc}

\usepackage[utf8]{inputenc}
\DeclareUnicodeCharacter{2500}{\rule[0.5ex]{1em}{0.55pt}}

\usepackage{microtype}

\usepackage{inconsolata}

\usepackage{graphicx}
\usepackage{amsmath}
\usepackage{multirow}
\usepackage{subcaption}
\usepackage{amsfonts}
\newtheorem{definition}{Definition}
\usepackage{booktabs}
\usepackage[most]{tcolorbox}

\newcommand{\beftext}[1]{}
\newcommand{\eat}[1]{}

\usepackage{multirow}
\usepackage[table,xcdraw]{xcolor}

\title{DSWorld: A Data Science World Model for Efficient Autonomous Agents
}

\author{
  \textbf{Zherui Yang\textsuperscript{1}},
  \textbf{Fan Liu\textsuperscript{1}},
  \textbf{Hao Liu\textsuperscript{1}\thanks{Corresponding author.}}
\\
\\
  \textsuperscript{1}The Hong Kong University of Science and Technology (Guangzhou)
\\
  zyang582@connect.hkust-gz.edu.cn;
  fliu236@connect.hkust-gz.edu.cn;
  liuh@ust.hk
}

\begin{document}
\maketitle
\begin{abstract}
Despite strong capabilities in data understanding and decision-making, autonomous data science agents still heavily rely on trial-and-error workflows that involve expensive computation. This bottleneck motivates models that can anticipate the effects of data science operations before real execution. In this paper, we introduce the concept of \emph{Data Science World Model}, which model the data science execution environment by predicting environment state transitions conditioned on current workflow states and candidate operations. We further propose \textbf{DSWorld}, a practical framework that combines structured state construction, cost-aware routing, lightweight real execution, and an LLM-based simulator for expensive operations. To support training, we construct an 8K-scale transition trajectory dataset and introduce Reflective World Model Optimization, an error-aware reinforcement learning strategy for improving transition prediction. Experiments show that DSWorld accelerates RL-based agent training by approximately $14\times$ and search-based inference by approximately $3$-$6\times$ while maintaining competitive performance, and outperforms the strongest LLM baseline by 35.6\% on transition prediction tasks. The code is available at \url{https://anonymous.4open.science/r/DSWorld}.

\eat{
Developing a data science world model that simulates environment state transitions given code-based actions is crucial for reducing the heavy reliance of data science workflows and experiments on costly real-world code execution. To this end, we propose \textbf{DSWorld}, the first data science world model. Specifically, DSWorld leverages an LLM-based simulator to predict the next environment state given the current state and a code-based action. To support effective training, we construct an 8K-scale dataset combining real agent trajectories and synthetic data, and further propose Reflective World Model Optimization, an error-aware reinforcement learning framework that improves next-state prediction through reflective optimization. Extensive experiments demonstrate that DSWorld outperforms strong baselines by 35.6\% on average across multiple prediction tasks. In addition, DSWorld substantially improves the efficiency of autonomous data science agents, accelerating agentic RL training by approximately 14$\times$ and inference by approximately 3--6$\times$ while maintaining strong agent performance. The code is available at \url{https://anonymous.4open.science/r/DSWorld}.}
\end{abstract}

\section{Introduction}

Autonomous data science agents have recently been proposed to automate a wide range of data science tasks, ranging from exploratory data analysis to predictive modeling~\cite{DBLP:journals/corr/abs-2509-23988, DBLP:journals/corr/abs-2510-23587}. For example, ML-Master~2~\cite{DBLP:journals/corr/abs-2601-10402} achieves medal-level performance on 56.4\% of Kaggle competition tasks on MLE-Bench~\cite{DBLP:conf/iclr/ChanCJASMSLMPMW25}. Existing methods typically leverage test-time scaling strategies, exploring numerous candidate solutions through iterative trial-and-error workflows~\cite{DBLP:journals/corr/abs-2502-13138, DBLP:journals/corr/abs-2510-08511}. However, these strategies heavily rely on expensive analytical computation, including data processing, model training, evaluation, and workflow updating. As a result, the majority of execution time is spent on computation rather than agent reasoning. For example, ML-Master~\cite{DBLP:journals/corr/abs-2506-16499} spends over 86\% of its execution time on model training in MLE-Bench. Such heavy computational overhead substantially limits the efficiency and scalability of autonomous data science systems.

\begin{figure}[!t]
  \centering
  \begin{subfigure}{0.47\textwidth}
    \includegraphics[width=\linewidth]{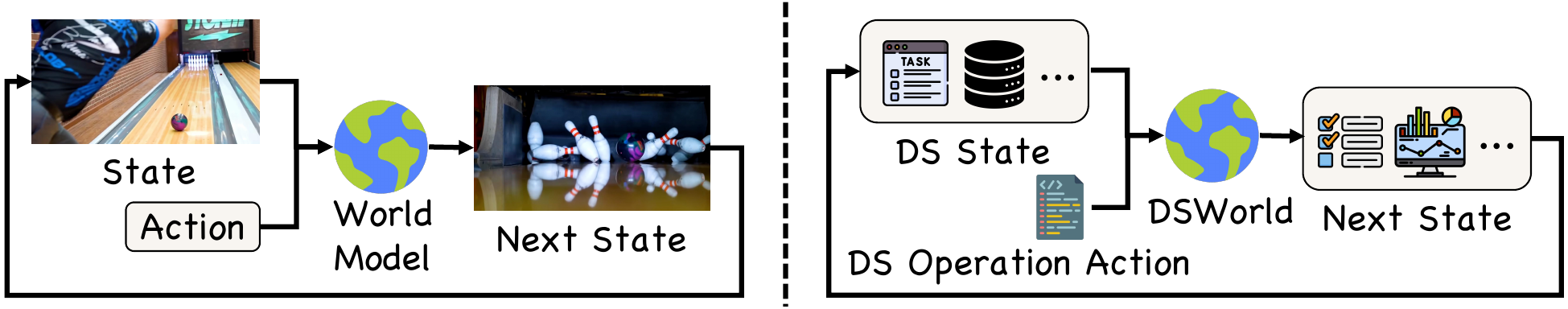}
    \caption{\textbf{Vision World Model} imagines future states of the physical world, while
\textbf{Data Science World Model} predicts the effects of data science operations without costly execution.}
    \label{fig:world}
  \end{subfigure}
  \begin{subfigure}{0.235\textwidth}
    \includegraphics[width=\linewidth]{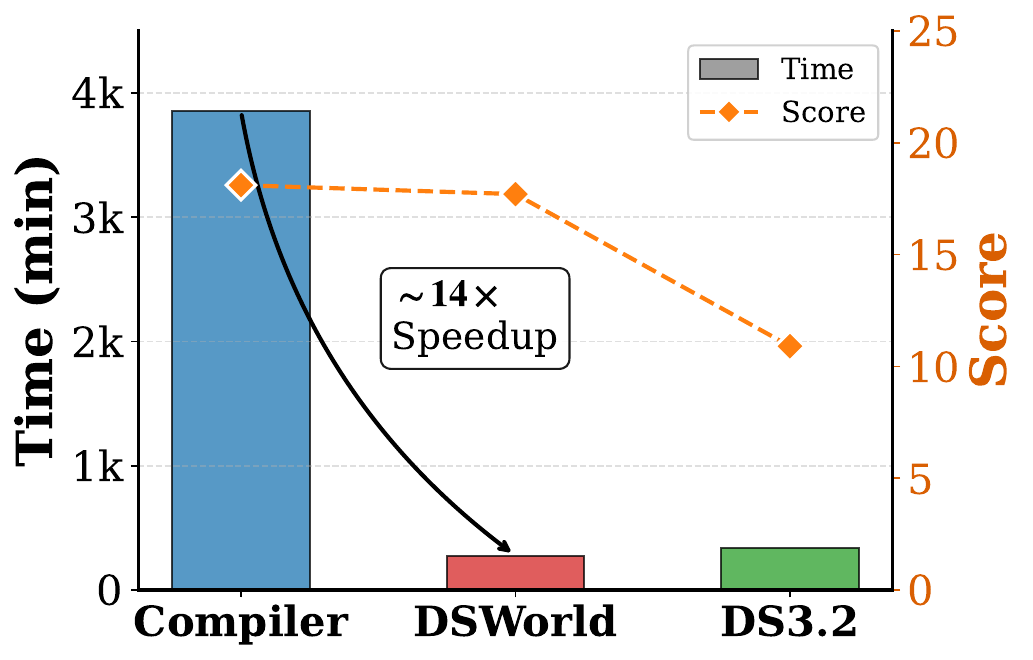}
    \caption{DSWorld accelerates agent RL training by $\sim14\times$.}
    \label{fig:rl}
  \end{subfigure}
  \begin{subfigure}{0.235\textwidth}
    \includegraphics[width=\linewidth]{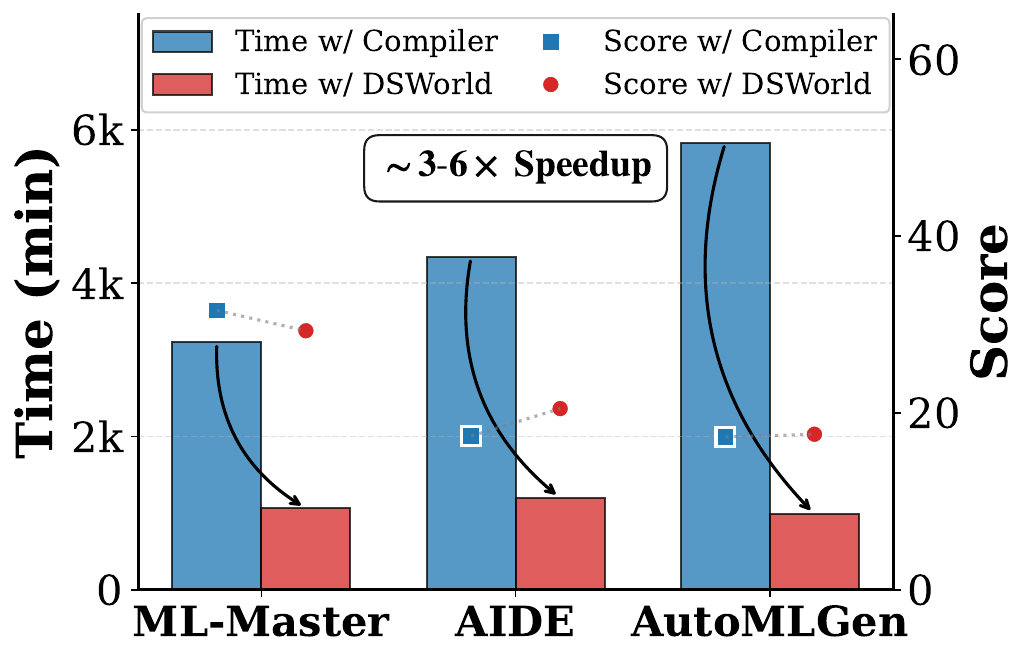}
    \caption{DSWorld accelerates agent inference by $\sim3$\text{-}$6\times$.}
    \label{fig:context_curve}
  \end{subfigure}
  \caption{DSWorld simulates data science environments and accelerates agent training and inference.}
  \label{fig:speedup}
\end{figure}

This raises a critical question: \emph{Can we develop a transition prediction model for data science workflows, enabling agents to anticipate the effects of data science operations before performing costly computation?} To this end, we introduce the concept of \emph{Data Science World Model}. As illustrated in Figure~\ref{fig:world}, similar to vision world models that imagine future states of the physical world~\cite{DBLP:journals/corr/abs-2604-22748, DBLP:journals/csur/DingZSZZFYSLSXL26}, data science world models treat the data science execution environment as the world to be modeled. Given a workflow state and a candidate data science operation, the model predicts the next environment state, including dataset and model changes, execution feedback, errors, and performance signals. Such a capability enables agents to anticipate the effects of data science operations without performing expensive real-world execution, thereby substantially accelerating both autonomous agent training and inference.

To realize this vision, we propose \textbf{DSWorld}, a practical data science world model framework for autonomous data science workflows. Specifically, a \emph{State Constructor} transforms raw data science environments into structured state representations containing tasks, datasets, execution histories, outputs, and environment status. To balance prediction accuracy and efficiency, we further introduce a cost-aware routing mechanism. Given an agent action, a \emph{Router} determines whether the action requires heavy computation. Lightweight actions are directly executed through a \emph{Compiler} to ensure accuracy, while computationally expensive actions are handled by an LLM-based \emph{Simulator}, which predicts the next environment state without actual execution to improve efficiency. Building upon this framework, we first perform supervised fine-tuning (SFT) for initialization and further propose \emph{Reflective World Model Optimization}, a reflective reinforcement learning strategy that improves next-state prediction through error-aware reflection and iterative refinement.

Due to the lack of large-scale transition data for autonomous data science workflows, we construct DSWorld-8K, a training dataset containing 8 thousand high quality data science agent trajectories. Specifically, we first collect authentic transition trajectories from real-world autonomous data science tasks and further synthesize corresponding Chain-of-Thought (CoT) trajectories~\cite{DBLP:conf/nips/Wei0SBIXCLZ22} to explain transition logic. However, real-world trajectories are expensive and inherently limited in scale. Therefore, we further develop a scalable data synthesis pipeline that leverages advanced LLMs and large-scale data sources from MMTU~\cite{DBLP:journals/corr/abs-2506-05587} to synthesize diverse workflow states and data science operations. The generated actions are executed in real environments to obtain next states, and only verified samples are retained for training. Additional CoT trajectories are synthesized to further explain transition dynamics.

We evaluate DSWorld in two practical downstream settings: (1) as a transition model for training autonomous data science agents, and (2) as a transition model for agent inference. Experimental results demonstrate that DSWorld substantially accelerates both agent training and inference while maintaining strong agent performance, achieving approximately $14\times$ RL training speedup and $3$-$6\times$ inference acceleration. In addition, we evaluate DSWorld across multiple transition prediction tasks. Experimental results show that DSWorld achieves strong predictive performance and outperforms the strongest LLM baseline by 35.6\% on average on transition prediction tasks.

In summary, our contributions are as follows:
\begin{itemize}
\item We introduce the concept of \emph{Data Science World Models}, which aim to predict the effects of data science operations in autonomous data science workflows without performing costly real-world computation.

\item We propose \textbf{DSWorld}, a practical data science world model framework for modeling state transitions, and further introduce a reflective reinforcement learning strategy that improves transition prediction through error-aware reflection and iterative refinement.

\item We develop a scalable data synthesis pipeline for constructing large-scale transition data for data science world model training.

\item Extensive experiments demonstrate the effectiveness of DSWorld, achieving approximately $14\times$ RL training speedup and $3$-$6\times$ inference acceleration while maintaining strong downstream agent performance.
\end{itemize}

\begin{figure*}[!t]
  \centerline{\includegraphics[width=\linewidth]{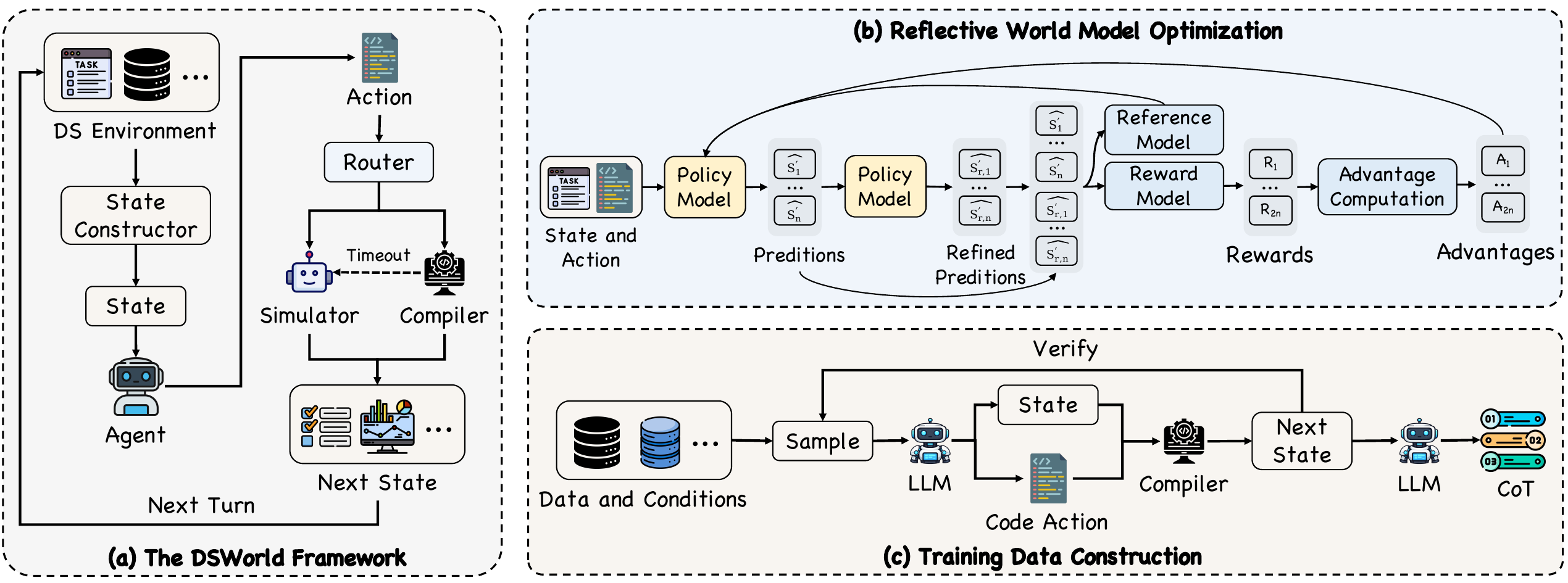}}
  \caption{Overview of DSWorld. (a) DSWorld predicts the effects of data science operations before performing costly computation. (b) Reflective World Model Optimization refines next-state prediction through error-aware reflection. (c) LLMs synthesize and verify data science state transitions and generate CoT trajectories.}
  \label{fig:framework}
\end{figure*}

\section{Related works}
\subsection{World Models}

World models aim to model environment dynamics for prediction, control, and simulation, typically by predicting future states conditioned on current states and actions~\cite{DBLP:journals/corr/abs-2604-22748, DBLP:journals/csur/DingZSZZFYSLSXL26, DBLP:conf/nips/HaS18}. Existing world model research spans multiple domains. In the Physical World, world models are used for video generation~\cite{DBLP:conf/icml/BruceDEPS0LMSAA24, chen2026learning}, 3D world generation~\cite{DBLP:journals/corr/abs-2509-07996, worldlabs2025marble}, and latent-space dynamics modeling~\cite{DBLP:conf/nips/HaS18, DBLP:conf/cvpr/AssranDMBVRLB23}. In the Digital World, recent works simulate webpage or software state transitions for web and GUI agents~\cite{DBLP:journals/corr/abs-2602-14721, ICLR2025_a0054803}. Despite these advances, world modeling for autonomous data science environments remains largely unexplored. Different from prior work, we study data science world models that model environment transitions in data science workflows.

\subsection{Autonomous Data Science Agents}

Recently, autonomous data science agents have been proposed to automate a wide range of data science tasks~\cite{DBLP:journals/corr/abs-2509-23988, DBLP:journals/corr/abs-2510-23587}. Existing methods mainly improve agent performance in two ways. One line of work enhances the backbone model through training. For example, ML-Agent~\cite{DBLP:journals/corr/abs-2505-23723} trains agents with reinforcement learning for machine learning tasks. Another line of work improves performance through test-time scaling strategies. For example, AIDE~\cite{DBLP:journals/corr/abs-2502-13138}, ML-Master~\cite{DBLP:journals/corr/abs-2506-16499}, AutoMLGen~\cite{DBLP:journals/corr/abs-2510-08511}, and AutoMind~\cite{DBLP:journals/corr/abs-2506-10974} use search algorithms to explore candidate solutions and select the optimal one. Despite their effectiveness, these approaches heavily rely on time-consuming computation during both training and inference. Therefore, developing data science world models that can model environment transitions and enable agents to anticipate the effects of data science operations before costly computation is critical.

\section{Preliminaries}

\begin{definition}[\textbf{Data Science Workflow State}]
The data science workflow state at time step $t$ is represented as $S_t = \{T_t, D_t, P_t, L_t\}$, where $T_t$ denotes the task, $D_t$ denotes the data state (e.g., dataset statistics and previews), $P_t$ denotes the execution environment (e.g., libraries and runtime configurations), and $L_t$ denotes execution logs, intermediate outputs, and task progress.
\end{definition}

\begin{definition}[\textbf{Action}]
At time step $t$, a data science agent produces an action $A_t$ conditioned on the current state $S_t$, where $A_t$ denotes data science operations such as feature engineering, model training, and evaluation.
\end{definition}

\begin{definition}[\textbf{Data Science World Model}]
A Data Science World Model is defined as a transition model that predicts the effects of data science operations before costly computation:
\begin{align}
    S_{t+1} = \mathcal{W}(S_t, A_t),
\end{align}
where $\mathcal{W}$ denotes the data science world model.
\end{definition}

\section{Methodology}
Figure~\ref{fig:framework} illustrates the overall framework of DSWorld. Given a data science workflow state, a State Constructor converts it into a structured representation. Based on the current state, a data science agent generates an action. The action is encoded into a dense embedding and routed to either a Compiler for real execution or an LLM-based Simulator for next-state prediction.

\subsection{DSWorld}

The proposed data science world model consists of four components $\mathcal{W} = \{\mathcal{SC}, \mathcal{R}, \mathcal{C}, \mathcal{S}\}$, where $\mathcal{SC}$ denotes the State Constructor, $\mathcal{R}$ denotes the Router, $\mathcal{C}$ denotes the Compiler, and $\mathcal{S}$ denotes the LLM-based Simulator.

\textbf{State Constructor.}
We first introduce the State Constructor, which transforms the raw execution environment into a structured state representation, formulated as $S_t = \mathcal{SC}(E_t)$, where $E_t$ denotes the data science environment at time step $t$.

Specifically, the State Constructor is a rule-based program that extracts and organizes key information from the environment, including task descriptions, dataset statistics, data previews, execution environments, execution histories, intermediate outputs, and error messages. The resulting structured state representation enables DSWorld to model environment transitions in a unified manner.

\textbf{Router.}
Given the current state, a data science agent generates an action as $A_t = \pi(S_t)$, where $\pi$ denotes the agent policy. To enable efficient routing, we first encode the generated action into a dense embedding representation using an action encoder. The resulting action embedding, together with the current state, is then fed into the Router for decision making. The Router determines whether the generated action can be executed efficiently as $m_t = \mathcal{R}(S_t, A_t)$, where $m_t \in \{\texttt{execute}, \texttt{simulate}\}$.

Intuitively, lightweight operations such as simple data manipulation or environment inspection are routed to direct execution, while computationally expensive operations, such as large-scale model training, are routed to simulation.

\textbf{Compiler.}
If the action is inexpensive, the Compiler executes it directly:
\begin{align}
\hat{S}_{t+1} = \mathcal{C}(S_t, A_t).
\end{align}
The Compiler interacts with the actual execution environment and returns the resulting next state, including updated data states, execution outputs, and runtime feedback.

\textbf{Simulator.}
Otherwise, the Simulator predicts the next state without real execution:
\begin{align}
\hat{S}_{t+1} = \mathcal{S}(S_t, A_t).
\end{align}
The Simulator is an LLM-based transition model that predicts execution outcomes and potential errors directly from the current state and action, thereby avoiding expensive computation.

To improve robustness against routing errors, we further impose a time limit on Compiler execution. If execution exceeds the predefined timeout threshold, the action is redirected to the Simulator for transition prediction. Thus, the overall transition process can be formulated as:

\begin{small}
\begin{align}
    \hat{S}_{t+1} =
\begin{cases}
\mathcal{C}(S_t, A_t), & m_t = \texttt{execute}, \\
\mathcal{S}(S_t, A_t), & m_t = \texttt{simulate} \ \text{or Timeout}.
\end{cases}
\end{align}
\end{small}
Through this hybrid execution-simulation mechanism, DSWorld balances efficiency and accuracy, enabling scalable environment interaction for autonomous data science agents.

\subsection{World Model Optimization}

To effectively train DSWorld and improve transition prediction quality, we adopt a two-stage post-training strategy consisting of SFT followed by Reflective World Model Optimization, a reflective reinforcement learning strategy that explicitly analyzes prediction errors and refines subsequent predictions to provide higher-quality training signals.

\subsubsection{SFT Warm-Up}

We first perform supervised fine-tuning on data science transition trajectories to initialize the Simulator with basic transition modeling capabilities. The SFT objective is defined as:
\begin{align}
    \mathcal{L}_{\text{SFT}}
    =
    -\log \mathcal{S}_\theta(S^{'} \mid S, A),
\end{align}
where $S$, $A$, and $S^{'}$ denote the current state, action, and next state, respectively.

\subsubsection{Reflective World Model Optimization}
After initializing DSWorld through SFT, we further optimize it with RL. Given the current state and action, the Simulator first predicts the next state as $\hat{S}^{'} \sim \mathcal{S}_\theta(\cdot \mid S, A)$. We then compare the prediction with the ground-truth next state and generate reflection feedback:
\begin{align}
    f = \mathcal{S}_\theta(\hat{S}^{'}, S^{'}),
\end{align}
where $f$ identifies missing, incorrect, or inconsistent predictions. Conditioned on the reflection feedback, the Simulator refines its prediction as
\begin{align}
    \hat{S}^{'}_{r}=\mathcal{S}_\theta(S, A, f).
\end{align}

For each training sample, we perform $n$ rollouts to obtain both original and refined predictions as $\mathcal{P} = \{\hat{S}_{i}^{'}, \hat{S}_{r,i}^{'}\}_{i=1}^{n}$. We jointly optimize all trajectories using Group Relative Policy Optimization (GRPO)~\cite{DBLP:journals/corr/abs-2402-03300}. The overall objective is defined as:

{\small
\begin{equation}
\begin{aligned}
\mathcal{L}(\theta) = &
\mathbb{E} \Bigg[
\frac{1}{2n} \sum_{i=1}^{n}\Big(
\mathcal{L}_{\text{clip}}(\hat{S}^{'}_i, A_i)
+
\mathcal{L}_{\text{clip}}(\hat{S}^{'}_{r,i}, A_{r,i})
\Big)\Bigg]
\\
&-
\beta_{\text{KL}} \, \mathbb{D}_{\text{KL}}(\pi_\theta \,\|\, \pi_{\text{ref}}),
\end{aligned}
\end{equation}
}
where
{\small
\begin{equation}
\begin{aligned}
\mathcal{L}_{\text{clip}}(\hat{S}^{'}, A) = &
\frac{1}{|\hat{S}^{'}|} \sum_{t=1}^{|\hat{S}^{'}|}
\min \Bigg[
\frac{\pi_\theta(y_t|x, y_{<t})}
{\pi_{\theta_{\text{old}}}(y_t|x, y_{<t})}
\mathcal{A},
\\
&
\text{clip}\!\left(
\frac{\pi_\theta(y_t|x, y_{<t})}
{\pi_{\theta_{\text{old}}}(y_t|x, y_{<t})},
1-\varepsilon,
1+\varepsilon
\right)
\mathcal{A}
\Bigg].
\end{aligned}
\end{equation}
}

Here, $\varepsilon$ and $\beta_{\text{KL}}$ are hyperparameters, $\pi_{\text{ref}}$ denotes a reference policy, and $\mathbb{D}_{\text{KL}}$ denotes the Kullback--Leibler divergence. The advantage $\mathcal{A}$ is computed as a group-relative advantage:
\begin{align}
\mathcal{A}_i =
\frac{R_i - \mu(R)}{\sigma(R) + \epsilon},
\end{align}
where $R_i$ denotes the reward of the $i$-th rollout, $\mu(R)$ and $\sigma(R)$ denote the mean and standard deviation of rewards within the rollout group, and $\epsilon$ is a small constant for numerical stability. The reward function evaluates whether the Simulator correctly predicts execution status, outputs, error information, and task performance.

\subsection{Training Data Construction}

Due to the lack of state transition data for data science workflows, we construct a high-quality training dataset, DSWorld-8K, consisting of both real and synthesized state transitions. To support transition reasoning learning, we additionally synthesize reasoning explanations for each transition. Each sample is represented as $(S, A, S^{'}, \tau)$, where $S$ is the current state, $A$ is the action, $S^{'}$ is the next state, and $\tau$ is the reasoning trajectory explaining the transition process.

\subsubsection{Real-World Transition Collection}

We first leverage existing autonomous data science agents and real-world data science benchmarks to collect authentic environment transition trajectories. Specifically, we run agents on real-world tasks and record the resulting environment transitions as $(S, A, S^{'})$. Based on the collected transitions, we further prompt an advanced LLM to synthesize corresponding CoT reasoning trajectories:
\begin{align}
\tau = p_\eta(S, A, S^{'}),
\end{align}
where $p_\eta$ denotes the LLM.

\subsubsection{Synthetic Transition Construction}

However, collecting real-world transition data is expensive and inherently limited in scale. To enable scalable training, we further design a synthetic pipeline for generating transition trajectories.

\textbf{State Synthesis.}
To construct diverse environment states, we leverage datasets from MMTU~\cite{DBLP:journals/corr/abs-2506-05587}, which contains over 60K real tables across diverse domains, providing rich and heterogeneous data science environments. For each sample, we randomly select a dataset and construct the corresponding state using the State Constructor as $S = \mathcal{SC}(E)$, where $E$ denotes the sampled data science environment. The constructed state includes task descriptions, dataset statistics, and data previews.

\textbf{Action Synthesis.}
To generate diverse actions, we construct data operation and execution error libraries based on the NumPy and Pandas ecosystems. Each entry contains an operation or error type, textual descriptions, and code examples. For each synthesized sample, we randomly sample a data operation $o$, an error type $e$, and an execution status $r \in \{\texttt{success}, \texttt{failure}\}$. Conditioned on the current state and sampled attributes, the LLM synthesizes executable actions:
\begin{align}
    A \sim p_\eta(A \mid S, o, e, r),
\end{align}
where $A$ denotes the generated action. This strategy enables the training data to cover both successful executions and diverse execution failures in data science workflows.

\begin{table*}[!t]
\caption{Performance comparison on data science transition prediction tasks.}
\resizebox{\linewidth}{!}{
\label{tab:result}
\begin{tabular}{cc|cccccc|c}
\toprule
\multicolumn{2}{c|}{\multirow{2}{*}{Methods}} & \multicolumn{4}{c}{Execution Prediction}                                                                                                                 & \multicolumn{2}{c|}{Performance Prediction}                                 & \multirow{2}{*}{AVG. $\uparrow$} \\ \cmidrule(l){3-6} \cmidrule(l){7-8}
\multicolumn{2}{l|}{}  & ESP $\uparrow$                       & ETP $\uparrow$                       & ERS $\uparrow$                       & EKM $\uparrow$                       & PP $\uparrow$                        & PR $\uparrow$                        &                                  \\ \midrule
\multirow{5}{*}{Training-free} & Llama-3.1-8B      & 0.480±0.112 & 0.322±0.047 & 0.318±0.061 & 0.043±0.021 & 0.622±0.02  & 0.492±0.021 & 0.379 \\
& Qwen3-8B                                      & 0.710±0.009                          & 0.573±0.023                          & 0.508±0.010                          & 0.193±0.020                          & 0.840±0.012                          & 0.507±0.009                          & 0.555                            \\
& DeepSeek-3.2                                  & 0.628±0.023                          & 0.403±0.028                          & 0.420±0.022                          & 0.250±0.010                          & \underline{0.851±0.005} & \textbf{0.539±0.011}                 & 0.516                            \\
& GPT-4o                                        & 0.712±0.018                          & 0.502±0.033                          & 0.472±0.020                          & 0.173±0.009                          & 0.757±0.010                          & 0.492±0.013                          & 0.518                            \\
& o4-mini                                       & 0.680±0.010                          & 0.585±0.023                          & 0.489±0.015                          & 0.382±0.031                          & 0.789±0.005                          & 0.514±0.028                          & 0.576                            \\ \midrule
\multirow{5}{*}{\begin{tabular}[c]{@{}c@{}}Trained on\\ DSWorld-8K\end{tabular}} & Llama-3.1-8B-sft  & 0.872±0.028 & 0.852±0.028 & 0.798±0.029 & 0.529±0.017 & 0.809±0.011 & 0.498±0.012 & 0.726 \\
& Llama-3.1-8B-grpo & 0.910±0.017 & 0.892±0.01  & 0.829±0.014 & 0.533±0.011 & 0.819±0.011 & 0.502±0.016 & 0.747 \\
& Qwen3-8B-sft                                  & 0.917±0.010                          & 0.885±0.015                          & 0.843±0.008                          & \underline{0.574±0.020} & 0.849±0.017                          & 0.509±0.011                          & 0.763                            \\
& Qwen3-8B-grpo                                 & \underline{0.937±0.013} & \underline{0.912±0.008} & \underline{0.859±0.010} & 0.556±0.005                          & 0.848±0.007                          & 0.513±0.007                          & \underline{0.771}   \\
& \textbf{DSWorld (Ours)}                                & \textbf{0.950±0.005}                 & \textbf{0.922±0.003}                 & \textbf{0.871±0.005}                 & \textbf{0.575±0.018}                 & \textbf{0.856±0.001}                 & \underline{0.518±0.008} & \textbf{0.781}                   \\ \bottomrule
\end{tabular}}
\end{table*}

\textbf{Ground-Truth Construction and Verification.}
After generating executable actions, we execute the synthesized action using the Compiler to obtain the corresponding next state:
\begin{align}
S^{'} = \mathcal{C}(S, A).
\end{align}

The resulting transition captures actual execution outcomes, including updated data states, execution outputs, runtime logs, and error messages. To ensure data quality, we further verify whether the execution results satisfy the intended execution status and error constraints:
\begin{align}
    \text{Verify}(S, A, S^{'}, e, r).
\end{align}

Only valid and consistent samples are retained. For each verified transition tuple $(S, A, S^{'})$, we further prompt the LLM to synthesize the corresponding CoT reasoning trajectory:
\begin{align}
    \tau = p_\eta(S, A, S^{'})
\end{align}
Combining both real-world and synthesized trajectories, we construct a final training dataset containing approximately 8K transition samples with corresponding reasoning trajectories.

\section{Experiments}
This section aims to answer the following research questions:
\textbf{RQ1}: How effective is DSWorld on data science transition prediction tasks?
\textbf{RQ2}: Can DSWorld effectively support the training and inference of agents?
\textbf{RQ3}: Is the proposed optimization strategy effective for training DSWorld?
\textbf{RQ4}: How do training data scale and model scale affect DSWorld performance?

\subsection{Experimental Setup}

\textbf{Benchmarks.}
We evaluate DSWorld on the Predict-before-Execute~\cite{DBLP:journals/corr/abs-2601-05930} benchmark to assess its performance ranking (\textbf{PR}) capability. In addition, since no benchmark exists for data science world models, we construct 540 evaluation tasks to measure execution success prediction (\textbf{ESP}), error type prediction (\textbf{ETP}), execution result similarity (\textbf{ERS}), execution keyword matching (\textbf{EKM}), and performance prediction (\textbf{PP}). Furthermore, we evaluate DSWorld as an environment simulator for autonomous data science agents on MLE-Bench Lite~\cite{DBLP:conf/iclr/ChanCJASMSLMPMW25}. Detailed benchmark descriptions are provided in Appendix~\ref{appendix:dataset}.

\textbf{Metrics.}
For PR, ESP, ETP, and EKM, we use accuracy as the evaluation metric. For ERS, we use embedding cosine similarity, while for PP, we use $1-\text{RMSE}$. For MLE-Bench Lite, we report the medal rate and overall score. Detailed metric descriptions are provided in Appendix~\ref{appendix:metric}.

\textbf{Baselines.}
Since no existing methods are specifically designed for data science world modeling, we compare DSWorld with two categories of baselines: (1) advanced LLMs, including Llama-3.1-8B~\cite{DBLP:journals/corr/abs-2407-21783}, Qwen3-8B~\cite{DBLP:journals/corr/abs-2505-09388}, DeepSeek 3.2~\cite{DBLP:journals/corr/abs-2512-02556}, GPT-4o~\cite{openai2023gpt4}, and o4-mini; and (2) trained LLM baselines, including Llama-3.1-8B-sft, Llama-3.1-8B-grpo, Qwen3-8B-sft, and Qwen3-8B-grpo.

\textbf{Implementation Details.}
DSWorld employs Qwen3-8B as the simulator backbone. The encoder is implemented using Harrier OSS v1 0.6B~\cite{microsoft2026harrier}, while the Router is implemented as a two-layer MLP. Detailed implementation settings are provided in the Appendix~\ref{appendix:implementation}.

\begin{table*}[!t]
\caption{Performance and training time comparison of agents trained with different environment simulators.}
\label{tab:agent_train} 
\centering
\resizebox{\linewidth}{!}{
\begin{tabular}{c|c|ccccc|cc}
\toprule
Backbones & Simulators            & Gold $\uparrow$     & Silver $\uparrow$    & Bronze $\uparrow$   & Any $\uparrow$       & Median $\uparrow$    & Score $\uparrow$    & Time (min) $\downarrow$      \\ \midrule
Qwen3-8B & -           & 1.59±2.75 & 1.59±2.75 & 1.59±2.75 & 4.76±4.76  & 7.94±2.75  & 13.80±0.53 & -\\
Qwen3-14B & -          & 1.59±2.75 & 0.00±0.00 & 6.35±2.75 & 7.94±2.75  & 7.94±2.75  & 16.50±2.75 & -\\ \hline
Qwen3-8B & DeepSeek 3.2	& 1.59±2.75	& 0.00±0.00	& 0.00±0.00	& 1.59±2.75	& 6.35±2.75	& 10.86±1.07 & 3854\\
Qwen3-8B & Compiler  & 1.59±2.75 & 1.59±2.75 & 6.35±2.75 & 11.11±2.75 & 12.70±2.75 & \textbf{18.11±1.93} & \underline{335}\\
Qwen3-8B & DSWorld & 1.59±2.75 & 1.59±2.75 & 6.35±2.75 & 9.52±0.00  & 11.11±2.75 & \underline{17.67±2.27} & \textbf{277}\\ \bottomrule
\end{tabular}}
\end{table*}

\subsection{Transition Prediction Performance (RQ1)}

Table~\ref{tab:result} presents the overall performance comparison between DSWorld and strong LLM baselines across multiple prediction tasks. Overall, DSWorld consistently achieves the best performance on nearly all evaluation dimensions, improving the average performance by 35.6\% over the strongest baseline, o4-mini. These results demonstrate the effectiveness of DSWorld for modeling data science environment transitions.

Specifically, compared with general-purpose LLMs, DSWorld substantially improves prediction accuracy on execution-related tasks. Compared with the strongest baseline, DSWorld achieves improvements of 33.4\%, 57.6\%, 71.5\%, and 50.5\% on these tasks, respectively. These results indicate that DSWorld can more accurately model execution dynamics and environment transitions in data science workflows. In addition, both SFT- and GRPO-trained models significantly outperform their untuned backbones, validating the effectiveness of the proposed synthetic transition data for training.

For performance-related tasks, DSWorld achieves competitive results with the best-performing methods. DSWorld achieves the best performance prediction score and the second-best performance ranking result. We attribute this to the fact that these tasks require stronger reasoning about machine learning algorithms, task characteristics, and evaluation metrics, making them more difficult to learn. In contrast, execution-related prediction tasks exhibit more explicit execution patterns and environment dynamics, which are easier for DSWorld to model effectively.

\begin{figure}[!t]
  \centerline{\includegraphics[width=0.9\linewidth]{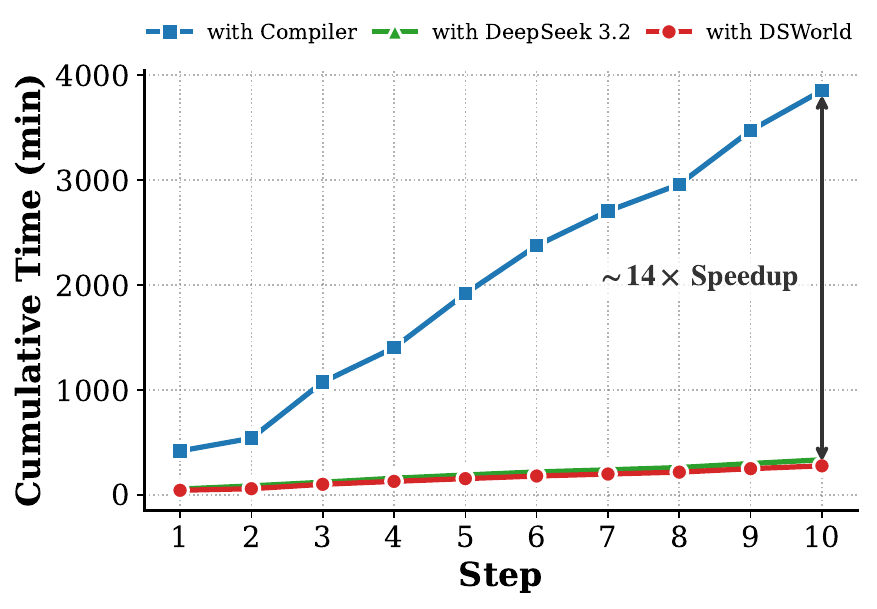}}
  \caption{RL training time across different simulators.}
  \label{fig:time_compare}
\end{figure}

\subsection{Training Agents with DSWorld (RQ2)}
\label{sec:trained_agent}

To investigate whether DSWorld can effectively support agent training, we use 105 machine learning tasks from MLE-Dojo~\cite{DBLP:journals/corr/abs-2505-07782} to train agents under the same ReAct~\cite{DBLP:conf/iclr/YaoZYDSN023} framework for 10 RL steps and evaluate them on MLE-Bench Lite. As shown in Table~\ref{tab:agent_train}, the agent trained with DSWorld achieves strong performance and remains competitive with Compiler-based training, while outperforming training with DeepSeek 3.2 as the simulator. Notably, the DSWorld-trained agent also outperforms the stronger Qwen3-14B baseline despite using Qwen3-8B as the backbone. These results demonstrate that DSWorld can serve as an effective training environment for autonomous data science agents.

More importantly, Figure~\ref{fig:time_compare} also compares the training efficiency of different simulators. Compiler-based training with real execution is significantly more time-consuming, whereas DSWorld achieves approximately $14\times$ acceleration while maintaining comparable downstream performance. Although Compiler-based training achieves slightly better final performance, the gap is relatively small compared with the substantial efficiency gain of DSWorld. These results further demonstrate the efficiency and scalability advantages of DSWorld.

\subsection{Agent Inference with DSWorld (RQ2)}
\begin{table*}[!t]
\caption{Performance and inference efficiency comparison using different executers on MLE-Bench Lite.}
\label{tab:search_agent}
\resizebox{\linewidth}{!}{
\begin{tabular}{c|c|c|ccccc|cc}
\toprule
Backbones                     & Methods                    & Executer     & Gold $\uparrow$                & Silver $\uparrow$              & Bronze $\uparrow$              & Any $\uparrow$                 & Median $\uparrow$               & Score $\uparrow$               & Time (s) $\downarrow$            \\ \midrule
\multirow{9}{*}{Qwen3-8B}     & \multirow{3}{*}{AIDE}      & Compiler     & 3.17±2.75            & 0.00±0.00            & 0.00±0.00            & 3.17±2.75            & 6.35±2.75             & 10.7±2.27            & 4102                 \\
                              &                            & DeepSeek 3.2 & 1.59±2.75            & 0.00±0.00            & 0.00±0.00            & 1.59±2.75            & 3.17±2.75             & 7.21±3.46            & 806                  \\
                              &                            & DSWorld      & 3.17±2.75            & 0.00±0.00            & 0.00±0.00            & 3.17±2.75            & 7.94±2.75             & 10.58±2.32           & 676                  \\ \cline{2-10} 
                              & \multirow{3}{*}{ML-Master} & Compiler     & 1.59±2.75            & 3.17±2.75            & 0.00±0.00            & 4.76±4.76            & 9.52±0.00             & 12.39±3.51           & 1421                 \\
                              &                            & DeepSeek 3.2 & 1.59±2.75            & 0.00±0.00            & 0.00±0.00            & 1.59±2.75            & 4.76±4.76             & 7.00±4.08            & 653                  \\
                              &                            & DSWorld      & 1.59±2.75            & 4.76±4.76            & 0.00±0.00            & 6.35±5.5             & 9.52±0.00             & 10.34±1.56           & 371                  \\ \cline{2-10} 
                              & \multirow{3}{*}{AutoMLGen} & Compiler     & 0.00±0.00            & 0.00±0.00            & 0.00±0.00            & 0.00±0.00            & 0.00±0.00             & 0.33±0.56            & 1470                 \\
                              &                            & DeepSeek 3.2 & 0.00±0.00            & 0.00±0.00            & 0.00±0.00            & 0.00±0.00            & 0.00±0.00             & 0.00±0.00            & 476                  \\
                              &                            & DSWorld      & 0.00±0.00            & 0.00±0.00            & 0.00±0.00            & 0.00±0.00            & 0.00±0.00             & 0.40±0.69            & 312                  \\ \midrule
\multirow{9}{*}{DeepSeek-3.2} & \multirow{3}{*}{AIDE}      & Compiler     & 4.76±4.76            & 3.17±2.75            & 1.59±2.75            & 9.52±4.76            & 14.29±4.76            & 17.41±4.13           & 4344                 \\
                              &                            & DeepSeek 3.2 & 3.17±2.75 & 3.17±5.50 & 4.76±4.76 & 11.11±5.50 & 14.29±9.52 & 15.12±5.04 & 1362 \\
                              &                            & DSWorld      & 6.35±2.75            & 3.17±5.50            & 3.17±5.50            & 12.70±9.91           & 17.46±7.27            & 20.49±3.27           & 1199                 \\ \cline{2-10} 
                              & \multirow{3}{*}{ML-Master} & Compiler     & 15.88±2.75           & 6.35±2.75            & 3.17±5.5             & 25.40±5.5            & 30.16±2.75            & 31.61±1.59           & 3232                 \\
                              &                            & DeepSeek 3.2 & 0.00±0.00            & 1.59±2.75            & 7.94±2.75            & 9.52±0.00            & 19.05±4.76            & 22.32±3.57           & 936                  \\
                              &                            & DSWorld      & 11.11±5.50           & 7.94±2.75            & 3.17±2.75            & 22.23±7.28           & 26.99±2.75            & 29.26±1.74           & 1065                 \\ \cline{2-10} 
                              & \multirow{3}{*}{AutoMLGen} & Compiler     & 3.17±2.75            & 0.00±0.00            & 3.17±5.5             & 6.35±7.27            & 15.87±7.27            & 17.29±4.36           & 5836                 \\
                              &                            & DeepSeek 3.2 & 0.00±0.00            & 0.00±0.00            & 3.17±2.75            & 3.17±2.75            & 9.52±2.75             & 12.46±2.63           & 1273                 \\
                              &                            & DSWorld      & 3.17±2.75            & 0.00±0.00            & 3.17±5.50            & 6.35±7.27            & 15.87±7.27            & 17.62±2.82           & 989                  \\ \bottomrule
\end{tabular}}
\end{table*}

We further investigate whether DSWorld can serve as an efficient inference environment for search-based autonomous data science agents. Specifically, we use the Compiler, DeepSeek 3.2, and DSWorld as the executers of search-based agents, including AIDE, ML-Master, and AutoMLGen. Table~\ref{tab:search_agent} presents the performance and efficiency results on MLE-Bench Lite. Overall, DSWorld consistently achieves competitive downstream performance while substantially reducing inference time across different agents and backbone models.

Specifically, compared with Compiler-based execution, DSWorld achieves approximately 3-6$\times$ acceleration while largely preserving downstream performance. In contrast, although directly using DeepSeek 3.2 also reduces execution time, it leads to severe performance degradation due to inaccurate execution feedback and hallucinated environment transitions. These results indicate that DSWorld can effectively simulate real environment transitions and provide sufficiently accurate execution feedback. Since search-based agents require evaluating many candidate solutions, replacing expensive real execution with efficient world model simulation can significantly reduce inference overhead while preserving search quality.

\eat{
\begin{figure}[!t]
  \centerline{\includegraphics[width=\linewidth]{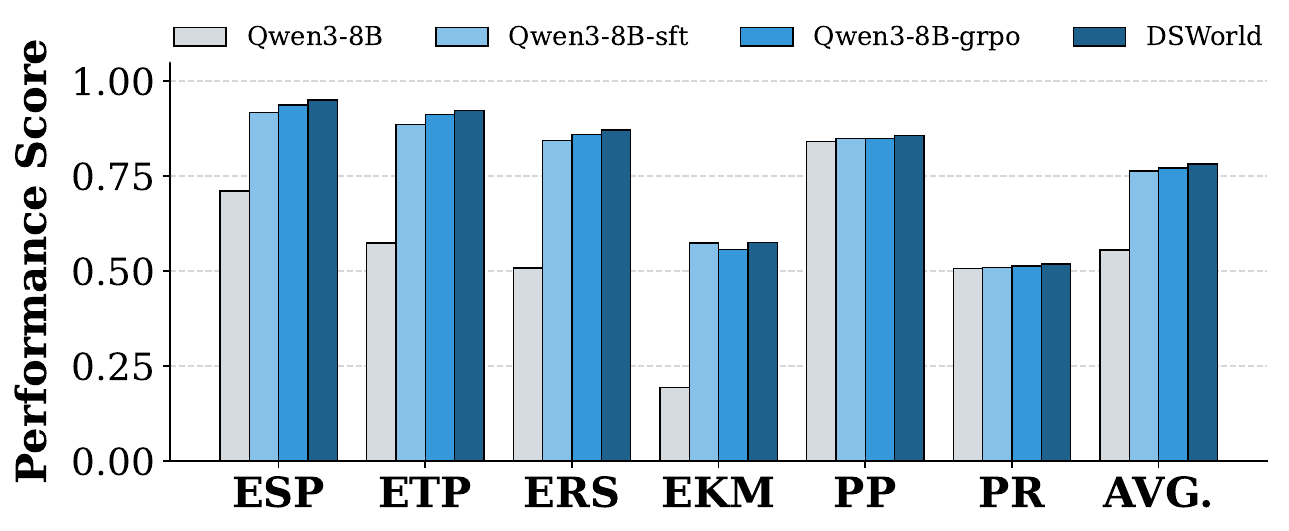}}
  \caption{Ablation study results.}
  \label{fig:ablation}
\end{figure}}

\subsection{Ablation Study (RQ3)}

To investigate the effectiveness of the proposed training strategy, we compare different variants of DSWorld. The results are shown in Table~\ref{tab:result}, where Qwen3-8B-sft and Qwen3-8B-grpo denote models trained with only SFT and with additional GRPO-based RL, respectively. Compared with the original Qwen3-8B backbone, Qwen3-8B-sft achieves substantial improvements across all tasks, improving the average performance by 37.5\%. These results demonstrate the effectiveness of the proposed data synthesis pipeline and supervised world model training. In particular, Qwen3-8B-sft significantly improves execution-related prediction tasks, indicating that supervised fine-tuning on large-scale transition trajectories enables the model to effectively learn data science environment dynamics.

Building upon Qwen3-8B-sft, Qwen3-8B-grpo further improves the overall performance by 1.05\%, demonstrating the effectiveness of RL for improving prediction quality through iterative refinement. Furthermore, DSWorld achieves the best overall performance across nearly all evaluation dimensions. Compared with Qwen3-8B-grpo, DSWorld further improves the overall performance by 1.3\%, and improves by 2.36\% over Qwen3-8B-sft, demonstrating the effectiveness of Reflective World Model Optimization. By introducing reflective error-aware optimization, DSWorld can iteratively refine transition reasoning and improve next-state prediction quality, enabling more accurate simulation of data science environment transitions.

\subsection{Further Analysis (RQ4)}
\begin{figure}[!t]
  \centering
  \begin{subfigure}{0.235\textwidth}
    \includegraphics[width=\linewidth]{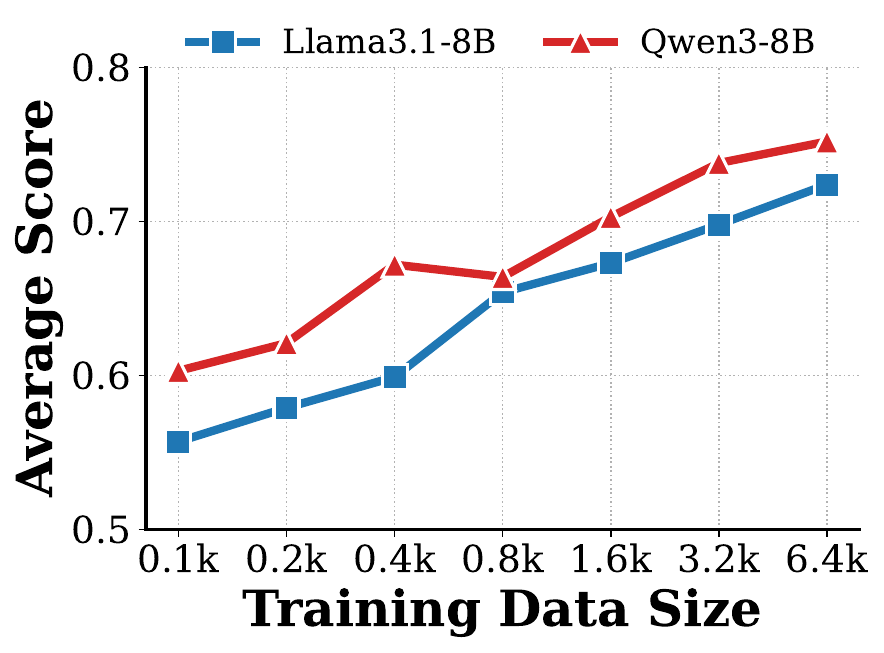}
    \caption{Performance under different training data scales.}
    \label{fig:context_curve}
  \end{subfigure}
  \begin{subfigure}{0.235\textwidth}
    \includegraphics[width=\linewidth]{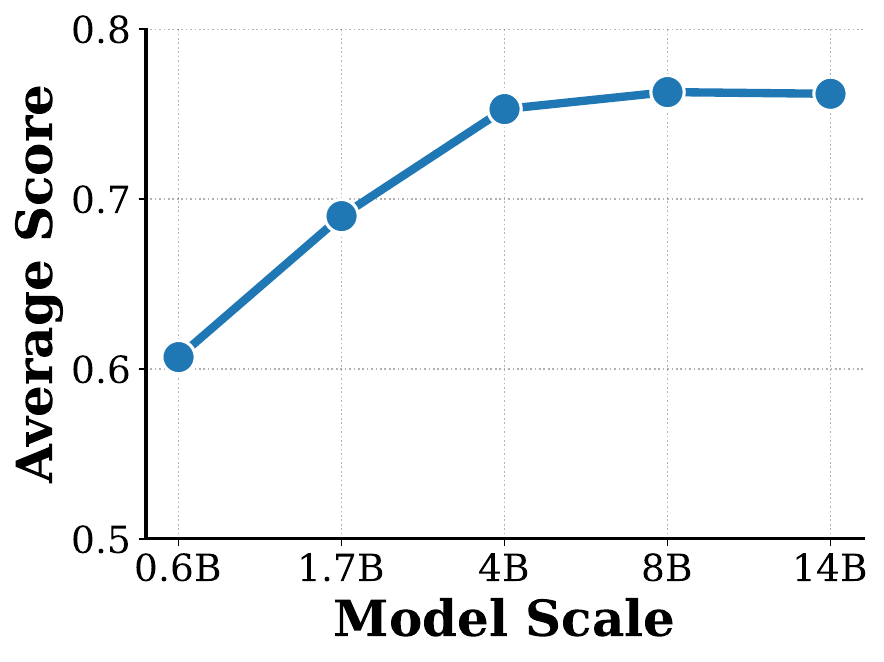}
    \caption{Performance under different Qwen3 model scales.}
    \label{fig:rl}
  \end{subfigure}
  \caption{Scale effects on DSWorld.}
  \label{fig:scale}
\end{figure}

In this section, we further investigate the effects of training data size and model scale on DSWorld. Figure~\ref{fig:scale} shows that DSWorld consistently benefits from more training data and larger backbone models. Specifically, as the training data size increases from 0.1k to 6.4k samples, the performance of both Llama3.1-8B and Qwen3-8B steadily improves, demonstrating the importance of large-scale transition data for world model learning. Moreover, scaling the Qwen3 backbone from 0.6B to 14B parameters also leads to significant performance gains, suggesting that stronger LLMs provide more accurate environment transition modeling and reasoning capabilities. Overall, these results indicate that DSWorld scales favorably with both data and model size, highlighting its potential for further improvement with larger-scale training resources.

\section{Conclusion}
In this paper, we present DSWorld, a learned transition model for data science workflows. By modeling environment transitions conditioned on the current state and actions, DSWorld enables agents to anticipate the effects of data science operations before performing costly computation during both training and inference. To support effective world model learning, we construct a large-scale trajectory dataset combining real and synthetic transition data, and further propose Reflective World Model Optimization to improve prediction quality through reflective reinforcement learning. Extensive experiments demonstrate that DSWorld achieves strong predictive performance across diverse transition prediction tasks while substantially accelerating autonomous agent training and inference.

\section*{Limitations}
Despite the promising results, this work still has several limitations. First, DSWorld currently focuses on modeling data science transitions and does not explicitly model external tool-call transitions in autonomous agent workflows. Second, the quality of transition prediction remains constrained by the capability of the underlying LLM simulator, which may occasionally produce inaccurate predictions in complex workflow scenarios. Third, synthesized trajectories may still exhibit distribution gaps compared with real-world autonomous workflows, potentially affecting generalization performance. We leave these limitations for future work.


\bibliography{custom}

\appendix

\section{Experimental Setup}
\label{appendix:setting}

\subsection{Benchmarks}
\label{appendix:dataset}

\begin{itemize}

\item \textbf{Predict-before-Execute~\cite{DBLP:journals/corr/abs-2601-05930}.}
Predict-before-Execute is a performance ranking benchmark containing 28 machine learning problems and 18,438 ranking tasks in total. Each task consists of a machine learning problem together with several candidate code solutions, and the model is required to predict which solution achieves better performance without executing the code. Since evaluating all 18,438 tasks is computationally expensive, we randomly sample 20 performance ranking tasks for each machine learning problem, resulting in 471 evaluation tasks in total.

\item \textbf{Synthetic Evaluation Tasks.}
Since there are currently no dedicated benchmarks for data science world modeling, we construct 540 evaluation tasks using our training data construction pipeline. To avoid data leakage, we use additional data sources from DABench~\cite{DBLP:conf/icml/HuZWCM0WSXZCY0K24} and MLE-Dojo~\cite{DBLP:journals/corr/abs-2505-07782}, which are different from the data sources used for synthesized training transitions. The tasks evaluate five core prediction capabilities:

\textbf{Execution Success Prediction (ESP).}
Predict whether a code action can execute successfully under the current environment state.

\textbf{Error Type Prediction (ETP).}
Predict the execution error category for failed code actions, such as syntax, runtime, or data-related errors.
    
\textbf{Execution Result Similarity (ERS).}
Evaluate the semantic similarity between predicted execution outputs and ground-truth execution results.
    
\textbf{Execution Keyword Matching (EKM).}
We use DeepSeek 3.2 to extract several keywords from the ground-truth execution results and evaluate whether the predicted execution outputs contain these keywords.
    
\textbf{Performance Prediction (PP).}
Given a machine learning task and a code solution, predict the resulting task performance.

\item \textbf{MLE-Bench Lite~\cite{DBLP:conf/iclr/ChanCJASMSLMPMW25}.}
MLE-Bench is a machine learning benchmark that requires agents to autonomously solve real-world machine learning tasks. Due to the high computational cost, we use MLE-Bench Lite for evaluation and further remove one task with a dataset exceeding 100GB. The final evaluation set contains 21 tasks.

\end{itemize}

\subsection{Metrics}
\label{appendix:metric}

For PR, ESP, and ETP, we use accuracy as the evaluation metric:
\begin{align}
\text{Score} =
\frac{1}{N}
\sum_{i=1}^{N}
\mathbf{1}(\hat{y}_i = y_i),
\end{align}
where $\hat{y}_i$ and $y_i$ denote the predicted and ground-truth labels, respectively.

For EKM, we use keyword matching accuracy to measure the proportion of correctly predicted keywords:
\begin{align}
\text{Score} =
\frac{1}{N}
\sum_{i=1}^{N}
\frac{|K_i^{p} \cap K_i^{g}|}
{|K_i^{g}|},
\end{align}
where $K_i^{p}$ and $K_i^{g}$ denote the predicted and ground-truth keyword sets, respectively.

For ERS, we use embedding cosine similarity between predicted outputs and ground-truth execution results:
\begin{align}
\text{Score} =
\frac{\mathbf{e}_{p}^{\top}\mathbf{e}_{g}}
{\|\mathbf{e}_{p}\|\|\mathbf{e}_{g}\|},
\end{align}
where $\mathbf{e}_{p}$ and $\mathbf{e}_{g}$ denote the embeddings of the predicted and ground-truth execution outputs, respectively. The embeddings are extracted using Harrier OSS v1 0.6B.

For PP, we use $1-\text{RMSE}$:
\begin{align}
\text{Score} =
1 -
\sqrt{
\frac{1}{N}
\sum_{i=1}^{N}
(\hat{s}_i - s_i)^2
},
\end{align}
where $\hat{s}_i$ and $s_i$ denote the predicted and ground-truth task performance scores, respectively.

For MLE-Bench Lite, we follow the official evaluation protocol and report the percentages of Gold, Silver, Bronze, and Any medals, as well as the percentage of tasks achieving above-median leaderboard performance. In addition, following MLE-Dojo, we compute a normalized leaderboard score ratio and use $1-\text{ratio}$ as the final score metric.

\subsection{Implementation Details.}
\label{appendix:implementation}
DSWorld employs Qwen3-8B as the simulator backbone. The encoder is implemented using Harrier OSS v1 0.6B~\cite{microsoft2026harrier}, while the Router is implemented as a two-layer MLP with hidden dimensions of 256 and 64, trained on collected code-execution time pairs. Data synthesis is performed using DeepSeek 3.2. For real-world transition collection, we use tasks from DACode~\cite{DBLP:conf/emnlp/HuangLYZLWHHLZL24} together with the ReAct framework. For synthesized transitions, we use data sources from MMTU~\cite{DBLP:journals/corr/abs-2506-05587}, a large-scale multi-task table understanding and reasoning benchmark containing 28,136 table-centric questions over 61,763 real tables across 25 task categories, providing diverse data science environments and analytical workflows. For all experiments, each task is evaluated over three independent runs, and we report the mean and variance of the results.

\begin{table*}[!t]
\caption{Performance of DSWorld under different execution environments.}
\resizebox{\linewidth}{!}{
\label{tab:environment}
\begin{tabular}{c|cccccc|c}
\toprule
\multicolumn{1}{c|}{\multirow{2}{*}{Environment}} & \multicolumn{4}{c}{Execution Prediction}                          & \multicolumn{2}{c|}{Performance Prediction} & \multirow{2}{*}{AVG. $\uparrow$} \\ \cmidrule(l){2-5} \cmidrule(l){6-7}
\multicolumn{1}{c|}{}                             & ESP $\uparrow$ & ETP $\uparrow$ & ERS $\uparrow$ & EKM $\uparrow$ & PP $\uparrow$     & PR $\uparrow$           &                                  \\ \midrule
Ubuntu                                             & 0.950±0.005    & 0.922±0.003    & 0.871±0.005    & 0.575±0.018    & 0.856±0.001       & 0.518±0.008             & 0.781                            \\
CentOS                                             & 0.943±0.006    & 0.923±0.005    & 0.864±0.007    & 0.562±0.005    & 0.856±0.005       & 0.513±0.008             & 0.777                            \\
Windows                                            & 0.941±0.009    & 0.916±0.007    & 0.872±0.005    & 0.569±0.017    & 0.853±0.007       & 0.516±0.009    & 0.778                            \\ \bottomrule
\end{tabular}}
\end{table*}

For SFT, we train the model for 5 epochs with a batch size of 32 and a learning rate of $1\times10^{-5}$. For RL, the reward function for execution prediction tasks is defined as the average score across the four execution prediction objectives:
\begin{align}
R_{\text{exec}}
=
\frac{1}{4}(
R_{\text{ESP}}
+
R_{\text{ETP}}
+
R_{\text{ERS}}
+
R_{\text{EKM}}).
\end{align}
For performance prediction tasks, the reward function is defined as:
\begin{align}
R_{\text{perf}}
=
1 -
(\hat{s} - s)^2,
\end{align}
where $\hat{s}$ and $s$ denote the predicted and ground-truth performance scores, respectively.

We use a rollout size of 8 and a learning rate of $1\times10^{-6}$ for 200 training steps. The maximum response length is set to 16K tokens. All experiments are conducted using VeRL~\cite{DBLP:journals/corr/abs-2509-01055} on 4 NVIDIA A800 GPUs.

\section{Additional Experiments}
\label{appendix:agent_training}
\subsection{Cross-Environment Generalization}

In this section, we evaluate the robustness of DSWorld across different execution environments, including Ubuntu, CentOS, and Windows systems. As shown in Table~\ref{tab:environment}, DSWorld achieves consistently strong performance across all environments, with only minor variations in both execution prediction and performance prediction metrics. In particular, Ubuntu achieves the best overall average performance, while CentOS and Windows remain highly competitive. These results demonstrate that DSWorld generalizes well across heterogeneous operating environments and is not overly dependent on a specific execution platform.

\begin{table}[!t]
\centering
\caption{Performance comparison of agents trained with different simulators.}
\label{tab:dacode_agent}
\begin{tabular}{c|c|c}
\hline
\textbf{Backbone} & \textbf{Simulator} & \textbf{DACode $\uparrow$} \\ \hline
Qwen3-8B  & -            & 0.200$\pm$0.018      \\
Qwen3-14B & -            & 0.214$\pm$0.006      \\ \hline
Qwen3-8B  & DeepSeek 3.2 & 0.158$\pm$0.012      \\
Qwen3-8B  & Compiler     & 0.231$\pm$0.009      \\
Qwen3-8B  & DSWorld      & \textbf{0.232$\pm$0.010}      \\ \hline
\end{tabular}
\end{table}

\subsection{Evaluating DSWorld as a Training Environment on Additional Benchmark}
To further evaluate the effectiveness of DSWorld as a training environment, we conduct additional experiments on 100 machine learning tasks from DACode~\cite{DBLP:conf/emnlp/HuangLYZLWHHLZL24}. Specifically, we train autonomous data science agents under the same ReAct framework using different simulators, including DeepSeek 3.2, real execution through the Compiler, and DSWorld.

Table~\ref{tab:dacode_agent} presents the results. Overall, agents trained with DSWorld achieve the best average performance among all settings. Consistent with the findings in Section~\ref{sec:trained_agent}, DSWorld-trained agents outperform the stronger Qwen3-14B baseline while only using Qwen3-8B as the backbone model. Compared with training using DeepSeek 3.2 as the simulator, DSWorld substantially improves downstream task performance, demonstrating stronger environment transition modeling capabilities. In addition, DSWorld achieves performance comparable to Compiler-based training while avoiding expensive real execution during training. These results further demonstrate that DSWorld can serve as an effective and scalable training environment for autonomous data science agents.

\section{Task Examples}
\label{appendix:example}
In this section, we provide two representative examples from our constructed evaluation tasks. The first example evaluates execution-level transition prediction, where the model predicts detailed execution outputs generated by data analysis code. The second example evaluates performance prediction, where the model estimates the downstream task performance of a machine learning solution without actual execution.

\begin{tcolorbox}[
    colback=gray!3,
    colframe=black!60,
    title=\textbf{Example 1: Execution Prediction},
    breakable
]

\textbf{Task.}
Please preprocess and analyze Microsoft stock market data to compute volatility, log returns, VWAP, momentum indicators, risk-adjusted returns, trend signals, and cumulative growth metrics.

\vspace{0.5em}
\textbf{Data.}
|   Unnamed: 0 | Date      |   Open |   High |   Low |   Close |   Volume | \\
|-------------:|:----------|-------:|-------:|------:|--------:|---------:| \\
|            0 | 19-Jan-18 |  90.14 |  90.61 | 89.66 |   90    | 36875013 | \\
|            1 | 18-Jan-18 |  89.8  |  90.67 | 89.66 |   90.1  | 24159683 | \\
\textbf{... omit here ...} \\
|           81 | 22-Sep-17 |  73.99 |  74.51 | 73.85 |   74.41 | 14111365 | \\
|           82 | 21-Sep-17 |  75.11 |  75.24 | 74.11 |   74.21 | 19186140 | 

\vspace{0.5em}
\textbf{Action.}
import pandas as pd \\
import numpy as np \\
df = pd.read\_csv('microsoft.csv') \\
df['Date'] = pd.to\_datetime(df['Date'], format='\%d-\%b-\%y')  \\
\textbf{... omit here ...} \\
final\_result = df[['Date', 'Close', 'VWAP', 'Risk\_Adjusted\_Return']].tail(10) \\
print(final\_result) 

\vspace{0.5em}
\textbf{Next State.}
         Date Close VWAP  Risk\_Adjusted\_Return\\
73 2017-10-04  74.69  83.796359             -2.640024\\
74 2017-10-03  74.26  83.725570             -0.914520\\
75 2017-10-02  74.61  83.641401              0.685921 \\
76 2017-09-29  74.49  83.548067             -0.234191 \\
77 2017-09-28  73.87  83.485573             -1.188027 \\
78 2017-09-27  73.85  83.376069             -0.039253 \\
79 2017-09-26  73.26  83.270265             -1.140838 \\
80 2017-09-25  73.26  83.131894              0.000000 \\
81 2017-09-22  74.41  83.062009              1.608727 \\
82 2017-09-21  74.21  82.966613             -0.278535 
\end{tcolorbox}

\begin{tcolorbox}[
    colback=gray!3,
    colframe=black!60,
    title=\textbf{Example 2: Performance Prediction},
    breakable
]

\textbf{Task.}
Machine Learning task: 

\#\#\# Description

In this competition, participants are tasked with developing a model that predicts employee access needs based on historical data. \\ \textbf{... omit here ...} \\
Participants should focus on creating a robust model that minimizes manual access transactions as employee attributes change over time.

\vspace{0.5em}
\textbf{Data.}
"--- COMPREHENSIVE DATA REPORT ---\\

\#\# Data Directory Structure\\

|-- description.txt\\
|-- sample\_submission.csv\\
|-- test.csv\\
|-- train.csv\\

\#\# Data Preview \\

\textbf{... omit here ...}

[END Preview]

\vspace{0.5em}
\textbf{Action.}
import pandas as pd\\
import numpy as np\\
import lightgbm as lgb\\
\# Load data\\
train = pd.read\_csv('train.csv')\\
test = pd.read\_csv('test.csv')\\
\textbf{... omit here ...}\\
submission = pd.DataFrame({'id': test\_ids, 'ACTION': test\_pred})\\
submission.to\_csv('submission.csv', index=False)

\vspace{0.5em}
\textbf{Score.}
0.813
\end{tcolorbox}

\section{Case Studies}
\label{appendix:case}
In this section, we present a representative case study to illustrate the prediction capabilities of DSWorld on real-world machine learning tasks. The example shows that DSWorld can effectively predict both execution outcomes and downstream task performance without actual code execution.

Specifically, DSWorld correctly predicts that the generated solution can execute successfully and produce a valid submission file. In addition, DSWorld accurately estimates the performance, achieving a predicted score of 0.528 compared with the ground-truth score of 0.543. The generated reasoning trajectory further demonstrates that DSWorld can analyze multiple aspects of the machine learning pipeline, including dataset characteristics, preprocessing strategies, feature representations, model selection, and potential performance limitations. For example, DSWorld recognizes that the TF-IDF and Logistic Regression pipeline provides a strong text classification baseline while also identifying its limitations in handling complex linguistic phenomena such as negation and sarcasm. These results demonstrate that DSWorld can capture high-level relationships between code implementations and downstream task performance, which is important for efficient autonomous data science agent planning and search.

\begin{tcolorbox}[
    colback=gray!3,
    colframe=black!60,
    title=\textbf{Case of DSWorld},
    breakable
]

\textbf{Task.}
Machine Learning task: \\

\#\#\# Description\\

Dive into the world of sentiment analysis with the Rotten Tomatoes movie review dataset, a rich corpus originally collected by Pang and Lee. This competition invites participants to label phrases on a five-point sentiment scale: negative, somewhat negative, neutral, somewhat positive, and positive. The challenge lies in navigating obstacles such as sentence negation, sarcasm, terseness, and language ambiguity. Hosted by Kaggle, this competition is designed for the machine learning community to engage in fun and practice, inspired by the pioneering work of Socher et al. Explore the accompanying website for source code, a live demo, and an online interface to aid in model training: http://nlp.stanford.edu/sentiment/.\\

\#\#\# Evaluation\\

Submissions are evaluated on classification accuracy (the percent of labels that are predicted correctly) for every parsed phrase. The sentiment labels are:\\

0 - negative  \\
1 - somewhat negative  \\
2 - neutral  \\
3 - somewhat positive  \\
4 - positive  \\

\textbf{... omit here ...}

\vspace{0.5em}
\textbf{Data.}
"--- COMPREHENSIVE DATA REPORT ---\\

\#\# Data Directory Structure\\

|-- description.txt\\
|-- sample\_submission.csv\\
|-- test.csv\\
|-- train.csv\\

\#\# Data Preview \\

\textbf{... omit here ...}

[END Preview]

\vspace{0.5em}
\textbf{Action.}
import pandas as pd\\
from sklearn.feature\_extraction.text import TfidfVectorizer\\
from sklearn.linear\_model import LogisticRegression\\

\# Load data\\
train\_df = pd.read\_csv('train.tsv', sep='\t')\\
test\_df = pd.read\_csv('test.tsv', sep='\t')\\

\# Preprocess\\
X\_train\_text = train\_df['Phrase'].fillna('').str.lower()\\
y\_train = train\_df['Sentiment']\\
X\_test\_text = test\_df['Phrase'].fillna('').str.lower()\\

\# TF-IDF vectorization\\
vectorizer = TfidfVectorizer(max\_features=10000, ngram\_range=(1, 2), stop\_words='english')\\
X\_train\_vec = vectorizer.fit\_transform(X\_train\_text)\\
X\_test\_vec = vectorizer.transform(X\_test\_text)\\

\# Train logistic regression on full data\\
model = LogisticRegression(max\_iter=1000, random\_state=42, class\_weight='balanced')\\
model.fit(X\_train\_vec, y\_train)\\

\# Predict on test\\
test\_predictions = model.predict(X\_test\_vec)\\

\# Create submission\\
submission = pd.DataFrame({\\
    'PhraseId': test\_df['PhraseId'],\\
    'Sentiment': test\_predictions\\
})\\

\# Save\\
submission.to\_csv('submission.csv', index=False)

\vspace{0.5em}
\textbf{Ground Truth.}
0.543

\vspace{0.5em}
\textbf{Prediction of DSWorld.}

\textbf{Reasoning.} <think>First, I assess the code's execution feasibility. The code imports standard libraries (pandas, sklearn) that are widely available in data science environments. The syntax appears correct with no obvious typos or indentation errors. The data loading uses pd.read\_csv with sep='\t', which is appropriate for TSV files as indicated in the dataset description. The column names referenced ('Phrase', 'Sentiment', 'PhraseId') align with typical expectations for this dataset, though I note that the data preview shows errors loading the TSV files, which could indicate potential issues with file paths or formatting. However, the code structure itself is sound and likely to run without syntax or immediate runtime errors.\\

Next, I examine the preprocessing and modeling pipeline. The use of TF-IDF vectorization with max\_features=10000, ngram\_range=(1,2), and stop\_words='english' is a reasonable baseline for text classification. The LogisticRegression configuration with max\_iter=1000 and class\_weight='balanced' addresses potential class imbalance, which is relevant given the five-class sentiment task. The pipeline follows a standard supervised learning workflow: load data, preprocess text, vectorize, train model, predict, and save results. There are no apparent logical inconsistencies, such as data leakage or incorrect handling of train/test splits.\\

However, I must consider potential hidden issues. The dataset description mentions that phrases are parsed from sentences, and some phrases may be repeated. The code does not explicitly handle duplicate PhraseIds, but this is unlikely to cause execution failure. The evaluation metric is classification accuracy, and the submission format requires PhraseId and Sentiment columns, which the code correctly generates and saves. The submission file name matches the requirement ('submission.csv').\\

Given the above, the code is highly likely to execute successfully without errors, assuming the TSV files exist at the expected paths and are correctly formatted. The absence of error handling for missing files or columns is a minor risk, but the dataset description provides clear file names and structure.\\

Now, for performance estimation: The approach uses a simple but standard method (TF-IDF + Logistic Regression) that has been shown to achieve moderate performance on sentiment analysis tasks. However, several factors may limit the score: (1) The model does not incorporate advanced text preprocessing (e.g., removing punctuation, handling negations) or feature engineering (e.g., sentiment lexicons). (2) No hyperparameter tuning or cross-validation is performed. (3) The dataset includes challenges like negation and sarcasm, which TF-IDF features may not capture well. (4) The class weighting is applied but may not fully address underlying distribution shifts.\\

For a five-class classification task with balanced class weighting, a reasonable baseline accuracy might be in the range of 0.4-0.6, assuming random guessing would yield ~0.2 (if classes were highly imbalanced) to ~0.25 (if balanced). Given the simplicity of the pipeline and the inherent difficulty of the task, I expect the accuracy to be modest but above random. The lack of advanced techniques (e.g., deep learning, BERT) suggests the score will not be state-of-the-art, but the TF-IDF + Logistic Regression combination often achieves solid performance on clean text data. Considering the evaluation context (Kaggle competition), a score around 0.5-0.6 is plausible, with the upper bound limited by the model's inability to handle complex linguistic patterns effectively. The final estimated score would reflect this baseline performance, acknowledging both the method's strengths and its limitations.\\</think>

\textbf{Score.} 0.528

\vspace{0.5em}
\end{tcolorbox}

\end{document}